\documentclass[10pt,a4paper,twocolumn]{article}

\usepackage{graphicx} 

\usepackage{dcolumn}           
\newcolumntype{.}   {D{.}{.}{-1}} 
\newcolumntype{d}[1]{D{.}{.}{#1}} 
\newcolumntype{e}   {D{E}{E}{-1}} 
\newcolumntype{E}[1]{D{E}{E}{#1}} 

\usepackage{amsmath}
\usepackage{mathptmx}

\usepackage{amssymb}
\usepackage{amsthm}

\usepackage{float}

\usepackage{setspace}

\usepackage{sectsty}

\newcommand{\myFontSize}{\fontsize{10}{12}\selectfont}
\renewcommand{\normalsize}{\fontsize{10}{12}\selectfont}
\sectionfont{\myFontSize}       
\subsectionfont{\rm\myFontSize\itshape} 

\usepackage{titlesec}
\titlespacing*{\section}{0pt}{10pt}{0pt}
\titlespacing*{\subsection}{0pt}{10pt}{0pt}

\usepackage{secdot}
\sectiondot{subsection}
\sectionpunct{section}{. }    
\sectionpunct{subsection}{. } 

\usepackage{caption}
\usepackage{subcaption}
\usepackage{xcolor} 
\usepackage{fancyhdr} 
\usepackage{lastpage}
\renewcommand{\headrulewidth}{0pt} 

\usepackage[numbers]{natbib}

\usepackage{amsmath}

\DeclareMathOperator*{\minimise}{minimise}

\usepackage[shortcuts,acronym,nonumberlist]{glossaries}
\makeglossaries 
\newacronym{CDM}{CDM}{conjunction data message}
\newacronym{TCA}{TCA}{time of closest approach}
\newacronym{SSN}{SSN}{Space Surveillance Network}
\newacronym{O/O}{O/O}{owners/operators}
\newacronym{NHPP}{NHPP}{non-homogeneous Poisson process}
\newacronym{MAE}{MAE}{mean absolute error}
\newacronym{RMSE}{RMSE}{root mean squared error}
\newacronym{MCMC}{MCMC}{Markov chain Monte Carlo}
\newacronym{NUTS}{NUTS}{No-U-Turn Sampler}

\newacronym{LEO}{LEO}{low-Earth orbit}
\newacronym{ESA}{ESA}{European Space Agency}

\begin{document}

%
\twocolumn[
\begin{@twocolumnfalse}

\vspace{0pt}
\begin{center}
    
    \vspace{15pt}
    \textbf{Statistical Learning of Conjunction Data Messages Through a Bayesian Non-Homogeneous Poisson Process}
    
    \vspace{10pt}
    \textbf{
        \selectfont\fontsize{10}{0}\selectfont Marta~Guimarães~\textsuperscript{a,b*},~Cláudia~Soares~\textsuperscript{b},~Chiara Manfletti\textsuperscript{a} 
    }
\end{center}


\vspace{-10pt} 
\begin{flushleft}
    \textsuperscript{a}\textit{
    \fontfamily{ptm}\selectfont\fontsize{10}{12}\selectfont Neuraspace, Portugal}, \underline{\{marta.guimaraes, chiara.manfletti\}@neuraspace.com}
    \\
    \textsuperscript{b}\textit{
        \fontfamily{ptm}\selectfont\fontsize{10}{12}\selectfont FCT-UNL, Portugal},
        \underline{claudia.soares@fct.unl.pt}
    \\
    \textsuperscript{*}\fontfamily{ptm}\selectfont\fontsize{10}{12}\selectfont Corresponding Author  
\end{flushleft}

\begin{abstract}
Current approaches for collision avoidance and space traffic management face many challenges, mainly due to the continuous increase in the number of objects in orbit and the lack of scalable and automated solutions. To avoid catastrophic incidents, satellite owners/operators must be aware of their assets’ collision risk to decide whether a collision avoidance manoeuvre needs to be performed. This process is typically executed through the use of warnings issued in the form of \glspl{CDM} which contain information about the event, such as the expected \gls{TCA} and the probability of collision. Our previous work presented a statistical learning model that allowed us to answer two important questions: (1) Will any new conjunctions be issued in the next specified time interval? (2) When and with what uncertainty will the next \gls{CDM} arrive? However, the model was based on an empirical Bayes homogeneous Poisson process, which assumes that the arrival rates of \glspl{CDM} are constant over time. In fact, the rate at which the \glspl{CDM} are issued depends on the behaviour of the objects as well as on the screening process performed by third parties. Thus, in this work, we extend the previous study and propose a Bayesian non-homogeneous Poisson process implemented with high precision using a Probabilistic Programming Language to fully describe the underlying phenomena. We compare the proposed solution with a baseline model to demonstrate the added value of our approach. The results
show that this problem can be successfully modelled by our Bayesian non-homogeneous Poisson Process with greater accuracy, contributing to the development of automated collision avoidance systems and helping operators react timely but sparingly with satellite manoeuvres.

\noindent{{\bf Keywords:}} Conjunction Data Message, Non-Homogeneous Poisson Process, Statistical Learning, Space Debris, Space Traffic Management \\

\end{abstract}

\end{@twocolumnfalse}
]


\section{Introduction}

The rapid growth of objects in Earth's orbit and the associated risk of collision poses significant challenges for collision avoidance and space traffic management. With the increasing number of satellites and debris, ensuring the safety and sustainability of space operations has become a critical concern. The European Space Agency estimates that the real number of objects larger than 1 centimetre is likely over one million~\cite{ESA2023}. Collisions between these objects generate even more debris, setting off a chain reaction known as the Kessler syndrome~\cite{Kessler1978}. To prevent such catastrophic failures, satellite \gls{O/O} must have a comprehensive understanding of the collision risks their assets face. 

\begin{figure}[htb!]
    \centering
    \includegraphics[width=0.45\textwidth]{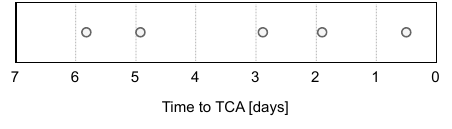}
    \caption{Example of CDMs' arrival times for a given conjunction in LEO. The grey circles represent a CDM.}
    \label{fig:conjunctionexample}
\end{figure}

To facilitate this monitoring process, the global \gls{SSN} detects, tracks, identifies, catalogues, and predicts the future states of space objects~\cite{NASA2020, Rongzhi2020}. By propagating the evolution of these objects' states over time, the \gls{SSN} assesses the likelihood of collisions. Each satellite, commonly referred to as the target, undergoes screening against all catalogued objects to identify potential close approaches, also known as conjunctions. Upon detection of a conjunction between the target and another object, often referred to as the chaser, the \gls{SSN} issues a \glsfirst{CDM}. These \glspl{CDM} provide essential information about the event, including the estimated \glsfirst{TCA} and the probability of a collision. Leading up to the \gls{TCA}, additional \glspl{CDM} are generated. These messages serve as timely warnings, allowing \gls{O/O} to assess the risks and determine the need for potential manoeuvring of their satellites. By continuously updating the information provided in the \glspl{CDM} as the \gls{TCA} is approached (Figure \ref{fig:conjunctionexample}), the \gls{SSN} enhances the accuracy and effectiveness of collision risk evaluation. However, due to the escalating number of space objects and the intricacies of their interactions, it is imperative to explore innovative approaches for managing collision risks.

\subsection{Related Work}

Predicting the arrival time of \glspl{CDM} is a relatively unexplored area of research, with only a limited number of studies addressing this specific problem.

The existing literature primarily focuses on space debris tracking, collision avoidance manoeuvres, and space traffic management. While these studies provide valuable insights into various aspects of space operations, very few works have specifically targeted the prediction of the next \gls{CDM} arrival time.

~\citet{Pinto2020} introduced a Bayesian deep learning approach to model all \gls{CDM} features simultaneously, including the time of arrival of future \glspl{CDM}, providing predictions of conjunction event evolution with associated uncertainties~\cite{Acciarini2020}. However, it is important to note that their study did not directly model the \gls{CDM} arrival process. Instead, their approach encompassed a broader scope by considering various aspects of conjunction events. 

Our previous work~\cite{Caldas2021}, specifically addresses the prediction of the next \gls{CDM} arrival time. We introduced a statistical learning model capable of answering two crucial questions: (i) Will any new conjunctions be issued in the next specified time interval? (ii) When and with what uncertainty will the next \gls{CDM} arrive? However, our previous model was based on an empirical Bayes homogeneous Poisson process, which assumes a constant arrival rate of \glspl{CDM} over time.

\subsection{Contributions}
As mentioned, the issuance rate of \glspl{CDM} is influenced by various factors, including the behaviour of space objects and the screening process. Consequently, to better represent the underlying phenomena, we extend our previous study and propose a Bayesian \gls{NHPP} implemented using a Probabilistic Programming Language, PyMC\footnote{Version 5.5.0.}~\cite{Salvatier2016}.

\section{Non-Homogeneous Poisson Process}

A \gls{NHPP} is a stochastic process that models the occurrence of events in continuous time. Unlike a homogeneous Poisson process, which assumes a constant event rate over time, a \gls{NHPP} allows for time-varying event rates. It provides a flexible framework for capturing complex temporal patterns and dynamics observed in real-world phenomena~\cite{Taylor1994, Last2017}.

Let $N\left( t \right)$ represent the \gls{NHPP}, where $t\ge 0$ denotes time. The process $N\left( t \right)$ satisfies the following properties:

\begin{enumerate}
    \item Increment Stationarity: The number of events in non-overlapping time intervals is independent. For any disjoint time intervals $\left[ s, t \right]$ and $\left[ u, v \right]$, where~$s < t \le u < v$, the random variables $N\left( t \right)-N\left( s \right)$ and $N\left( v \right)-N\left( u \right)$ are independent.

    \item Increment Independence: The number of events in a given time interval is independent of the number of events in non-overlapping time intervals. For any $0 \le s < t$, the random variables $N\left( t \right)-N\left( s \right)$ and $N\left( t + s\right)$ are independent.

    \item Time-Varying Event Rate: The intensity function $\lambda(t)$ determines the time-varying event rate at time $t$. It represents the instantaneous rate of events per unit time. Thus, the expected number of events in a short time interval is given by
    $$N\left( t + s\right) - N\left( t\right) \sim Poisson\left( \int_{t}^{t+s}\lambda\left( \tau \right)d\tau  \right) .$$
\end{enumerate}

\section{Problem Formulation}

\subsection{Intensity Function}

The intensity function $\lambda(t)$ can be deterministic or stochastic. In the case of a deterministic intensity function, the process follows a deterministic trend over time that can be used to add domain knowledge or to retrieve important information about expected behaviours. Stochastic intensity functions, on the other hand, introduce randomness in the event occurrence rates, allowing the process to capture more complex and unpredictable behaviours. With the cost of creating a deeper hierarchical model, we decided to use a deterministic intensity function. Firstly, while the behaviour of space objects and the screening process contribute to variations in the issuance rates of \glspl{CDM}, they can often be characterized by discernible patterns and trends. Thus, with a deterministic intensity function, we can capture and incorporate these known patterns into the model, allowing for a more accurate representation of the underlying dynamics. Furthermore, a deterministic intensity function offers computational advantages, as it simplifies the estimation process. This allows for faster model training and evaluation, which is crucial in real-time collision avoidance operations.

We opted to use a polynomial function family for the \gls{NHPP}. The choice of a polynomial function stems from its ability to capture a wide range of intensity patterns, including both linear and nonlinear trends. To ensure robustness and mitigate the potential overfitting issue, we employed polynomial Ridge regression~\cite{Murphy2022} to estimate the polynomial coefficients. Such a method incorporates a regularization term that helps prevent excessive sensitivity to noise in the training data. This regularization term adds a penalty to the model's objective function, promoting smoother and more stable solutions.

We denote the polynomial function family as
\begin{equation}
    \lambda(t) = \beta_0 + \beta_{1}t + \beta_{2}t^2 + ... + \beta_{m}t^m ,
    \label{eq:lambda_t}
\end{equation}
where $\beta_j$ are the coefficients of the polynomial terms. Thus, the objective function to be minimised is given by
\begin{equation}
    \minimise_{\beta} \quad \sum_{i}^{}\left[ N_i - \lambda\left( t_i \right) \cdot \Delta t_i \right]^2 + \alpha\sum_{j}^{} \beta_{j}^2 ,
    \label{eq:ridge}
\end{equation}
where $N_i$ represents the observed number of \glspl{CDM} at the $i$th time point and $\lambda\left( t_i \right) \cdot \Delta t_i$ denotes the predicted number of issued \glspl{CDM} using the polynomial function at the $i$th time point. Thus, the first term of \eqref{eq:ridge} represents the sum of squared errors or residuals between the observed number of issued \glspl{CDM} and the predictions made by the polynomial function at each time point. It measures the discrepancy between the actual and predicted \gls{CDM} counts, quantifying the goodness of fit of the polynomial function to the observed data.
The second term incorporates the regularization component into the objective function. It penalizes the magnitude of the coefficients, $\beta_j$, by adding the sum of their squared values multiplied by the hyperparameter $\alpha$. The $\alpha$ parameter determines the strength of regularization and controls the trade-off between fitting the data closely and controlling the complexity of the model.

By minimising this objective function, the Ridge regression aims to find the optimal values for the polynomial coefficients, $\beta_j$, that minimise the sum of squared errors while considering the regularization term. This approach allows for a balance between accurately modelling the observed \gls{CDM} counts and controlling the complexity of the polynomial function.

%

\subsection{Data Model}

To model a \gls{NHPP}, different techniques can be applied, such as empirical estimation or Bayesian methods. Empirical estimations involve analysing historical data to estimate the time-varying intensity function. While this approach provides a straightforward and data-driven estimation, it is limited by the availability and quality of historical data, particularly in rapidly evolving domains like space traffic management.
Bayesian methods, including Bayesian \glspl{NHPP}, present a robust and comprehensive approach for modelling the intensity function. By incorporating prior beliefs about the intensity function and updating them based on observed data, Bayesian inference offers a principled way to estimate the time-varying intensity. This approach allows for the integration of prior knowledge and quantification of uncertainty associated with the estimates through posterior distributions --- termed epistemic uncertainty.

\begin{figure}[htb!]
    \centering
    \includegraphics[width=0.22\textwidth]{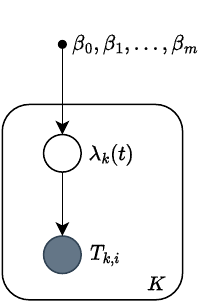}
    \caption{Graphical model of the proposed solution. $\beta_j$ are the coefficients of the polynomial terms, $\lambda_k\left( t \right)$ is the intensity function and $T_{k,i}$ is the time of arrival of the $i$th CDM of a given event $k$. }
    \label{fig:data_model}
\end{figure}

As indicated in Figure \ref{fig:data_model}, the parameters $\beta_j$, derived from the training set, represent valuable prior knowledge about the underlying behaviour of the intensity function. By incorporating this prior information, we can estimate the posterior distribution of each intensity function $\lambda_k\left( t \right)$, given a particular time interval.
Through Bayesian inference, we update our initial beliefs (represented by the prior distribution) about the intensity function based on the observed data. The prior distribution, informed by the parameters $\beta_j$, is combined with the likelihood of the observed \gls{CDM} counts to yield the posterior distribution of the intensity function for each event.
By obtaining the posterior distribution, we understand the uncertainty associated with the estimates of the intensity function within the specified time interval of interest. Once the intensity function is determined, we can use this information to predict the timing of the next \gls{CDM} occurrence, $T_{i}$.

\section{Predicting the Next CDM Arrival Time}

In scenarios where a conjunction carries a significant risk of collision, satellite \gls{O/O} are tasked with making critical decisions regarding whether to perform a collision avoidance manoeuvre. To ensure the manoeuvre is executed safely and efficiently, \gls{O/O} must decide up to two days prior to the \gls{TCA}. At this point in time, the most recent \gls{CDM} issued, which is at least two days before the \gls{TCA}, represents the most reliable information available to the \gls{O/O}.

Considering this, we adopted a specific approach during our analysis: we partitioned the test data set, separating it at 2.5 days before the \gls{TCA}, and aimed to predict the subsequent \gls{CDM} arrival time. This 2.5-day cutoff serves to provide some slack in the prediction process, as the decision regarding collision avoidance manoeuvres should ideally be made at least 2 days prior to the \gls{TCA}. This partitioning strategy aims to replicate the decision scenario faced by \gls{O/O} in real-world operations. In such scenarios, the key question revolves around whether it is advantageous to await a new \gls{CDM} containing fresh and updated information about the conjunction, or if it is more prudent to make a decision promptly based on the available information at that moment.

\subsection{Naïve Baseline Solution}
\begin{figure}[htb!]
    \centering
    \includegraphics[width=0.45\textwidth]{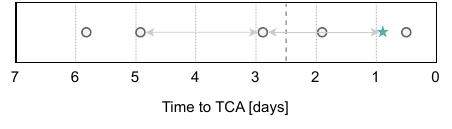}
    \caption{Example of the Baseline solution for a given conjunction in LEO when predicting 2.5 days before the TCA. The grey circles represent the real arrival times of CDMs, and the star indicates the prediction performed by the naïve Baseline model. The arrows indicate the predicted inter-CDM distance, equal to the previous one.}
    \label{fig:baseline}
\end{figure}

The naïve baseline solution we employ in our study reflects a common assumption made by satellite operators in real-life operations. When faced with the decision of whether to await a new \gls{CDM} or make a timely decision, operators often default to the notion that the previous inter-\gls{CDM} time serves as the best estimator for the next inter-\gls{CDM} time (Figure \ref{fig:baseline}). This assumption provides a practical rule of thumb that helps operators make timely decisions based on the available information without needing to account for more complex temporal dynamics.

\subsection{Mean-Based Baseline Solution}

In addition to the naïve Baseline solution discussed above, we consider a second Baseline approach that assumes the inter-\gls{CDM} time to be the mean of all the previous inter-\gls{CDM} times. This approach takes into account the historical inter-\gls{CDM} times of a given conjunction and calculates their average as a predictor for the next inter-\gls{CDM} time. Thus, this solution seeks to capture the central tendency or typical duration between consecutive \glspl{CDM}. 
It offers a more refined estimation of the inter-\gls{CDM} time compared to the naïve solution, which assumes a constant inter-\gls{CDM} time. By incorporating historical data, this approach acknowledges the potential variability in the issuance of \glspl{CDM} and attempts to provide a more informed prediction.


\begin{figure}[htb!]
    \centering
    \includegraphics[width=0.45\textwidth]{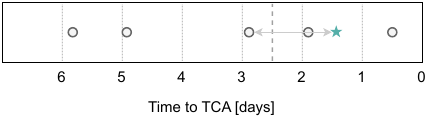}
    \caption{Example of the mean-based Baseline solution for a given conjunction in LEO when predicting 2.5 days before the TCA. The grey circles represent the real arrival times of CDMs, and the star indicates the prediction performed by the mean-based Baseline model. The arrow indicates the predicted inter-CDM distance, equal to the mean of all the previous ones.}
    \label{fig:mean_baseline}
\end{figure}



\subsection{NHPP Solution}

We propose a \gls{NHPP} solution to model the issuance of \glspl{CDM} over time. Two important aspects of our \gls{NHPP} model are the selection of the polynomial degree for the intensity function and the incorporation of prior distributions for the coefficients of the polynomial terms.

For our analysis, we focused on experimenting to model the expected value, $\Lambda(t) =\int_{T}\lambda\left( t \right)dt $, with a polynomial of degree three. The choice of degree three allows for capturing more complex temporal patterns and variations in the \gls{CDM} issuance rates compared to simpler linear or quadratic polynomials.

To fully account for uncertainties and incorporate prior knowledge about the coefficients of the polynomial terms, we modelled their prior distributions as Gaussian distributions. This approach allows us to consider various potential configurations of the polynomial and incorporate our prior beliefs about their distributions.

The corresponding mean, $\mu$, and standard deviation, $\sigma$, of such distributions were extracted from historical data and can be defined as
\begin{align}
    \beta_0\sim & \mathcal{N}(\mu=8.58, \sigma=3.42) \nonumber \\
    \beta_1\sim & \mathcal{N}(\mu=-0.54, \sigma=0.41) \nonumber \\
    \beta_2\sim & \mathcal{N}(\mu=-0.60, \sigma=0.37) \nonumber \\
    \beta_3\sim & \mathcal{N}(\mu=-0.01, \sigma=0.19) \nonumber .
\end{align}

Having defined the likelihood and prior distributions, it is possible to infer the posterior distributions, i.e., the parameters of the implemented approach. The determination of such parameters can be achieved by using the \gls{NUTS} algorithm~\cite{Homan2014}, which is readily accessible within the PyMC library.

The \gls{NUTS} is a highly efficient and widely used \gls{MCMC} algorithm designed for sampling from the posterior distribution of a target probability distribution~\cite{Neal1993}. \gls{MCMC} methods are used in Bayesian inference to approximate the posterior distribution of model parameters given observed data. The goal is to generate samples from this distribution. Traditional \gls{MCMC} algorithms suffer from slow convergence in high-dimensional and complex parameter spaces. \gls{NUTS} is specifically designed to address these limitations by leveraging concepts from Hamiltonian Monte Carlo, a more sophisticated \gls{MCMC} approach.

Throughout the sampling procedure, a total of four chains, each encompassing 1,000 iterations, were employed.

\begin{figure}[htb!]
    \centering
    \includegraphics[width=0.45\textwidth]{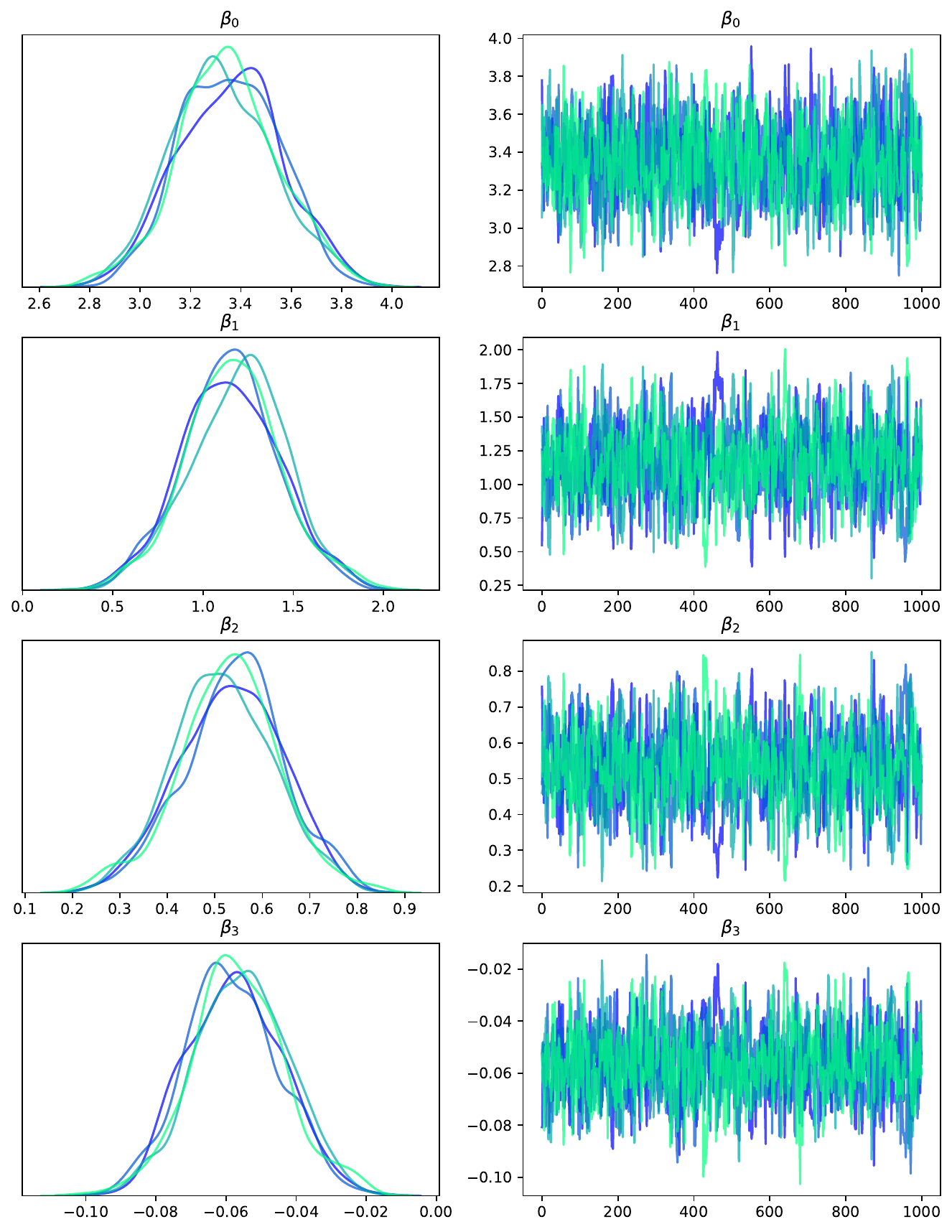}
    \caption{Posterior distributions of the coefficients of the terms defining the expected value $\Lambda(t)$ polynomial function. The left column consists of a smoothed histogram of the marginal posteriors of each random variable while the right column contains the samples of the Markov chain plotted in sequential order. The convergence of the chains is evident from the overlapping and consistent distributions.}
    \label{fig:posterior}
\end{figure}

Figure~\ref{fig:posterior} shows the obtained posterior distributions. As can be seen, the convergence of the chains is evident from the overlapping and consistent distributions, indicating successful exploration of the parameter space and reliable estimation of parameter uncertainty.


\subsection{Results}

Table \ref{tab:results} presents the results of our evaluation, comparing the proposed \gls{NHPP} model with the two baseline solutions: the naïve solution based on the previous inter-\gls{CDM} time as the predictor, and the baseline assuming the mean of all previous inter-\gls{CDM} times as the predictor.

\begin{table}[!htb]
\renewcommand{\arraystretch}{1.3}
\caption{MAE and RMSE obtained when predicting the next CDM arrival time at 2.5 days to the TCA.}
\label{tab:results}
\centering
\begin{tabular}{ccc}
\hline
\textbf{Model}        &  \textbf{MAE}    &  \textbf{RMSE}   \\
\textbf{}             &  \textbf{[days]} &  \textbf{[days]} \\ \hline
{Naïve Baseline}      & 0.199            &  0.318           \\
{Mean-Based Baseline}       & 1.039            &  1.087           \\
{NHPP}                & \textbf{0.124}   &  \textbf{0.191}             \\ \hline
\end{tabular}
\end{table}

The performance of the models was evaluated through the \gls{MAE} and \gls{RMSE}, defined as
\begin{equation*}
    MAE = \frac{\sum_{i=1}^{n}\left| y_{i} - \widehat{y}_{i} \right|}{n} ,
\end{equation*}
\begin{equation*}
    RMSE = \sqrt{\frac{\sum_{i=1}^{n}\left( y_{i} - \widehat{y}_{i} \right)^2}{n}} ,
\end{equation*}
where $n$ is the number of predicted samples, $y_{i}$ is the real arrival time and $\widehat{y}_{i}$ is the predicted arrival time. Both metrics are expressed in the same units as the variable being predicted, i.e., in days.

The results clearly demonstrate that the proposed \gls{NHPP} outperforms both baseline solutions, achieving significantly lower \gls{MAE} and \gls{RMSE} values.

\begin{figure}[!htb]
    \begin{subfigure}{.5\textwidth}
        \centering
        \includegraphics[width=\linewidth]{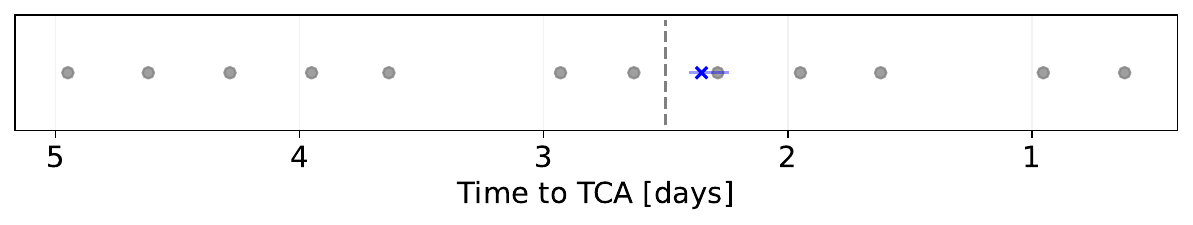}  
        \caption{Conjunction with a good prediction.}
        \label{fig:good_prediction}
        \end{subfigure}
        \begin{subfigure}{.5\textwidth}
        \centering
        \includegraphics[width=\linewidth]{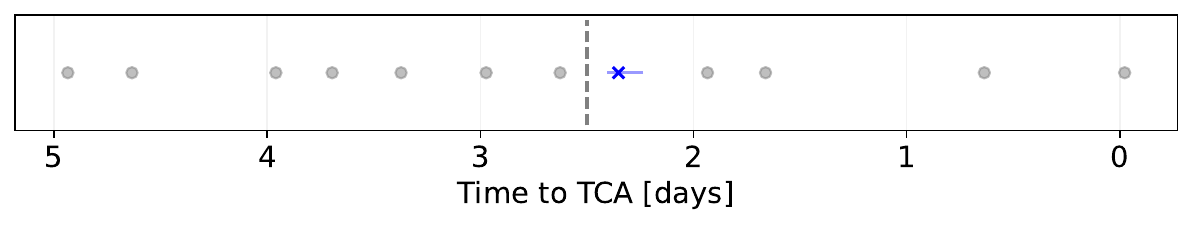}  
        \caption{Conjunction with a bad prediction.}
        \label{fig:bad_prediction}
    \end{subfigure}
    \caption{Example of the NHPP solution for two given conjunctions in LEO when predicting 2.5 days before the TCA. Figure~\ref{fig:good_prediction} showcases a good prediction while Figure~\ref{fig:bad_prediction} serves as an example of a bad prediction. The grey circles represent the real arrival times of CDMs, the cross indicates the prediction, and the blue line represents the 95\% uncertainty interval.}
    \label{fig:predictions}
\end{figure}

Figure~\ref{fig:predictions} shows two examples of the results obtained with the \gls{NHPP} model when predicting at 2.5 days prior to the \gls{TCA}. Each prediction, denoted by a blue cross, is associated with the correspondent 95\% estimated uncertainty, represented by the blue line.

The first example, Figure~\ref{fig:good_prediction}, shows that indeed the model can achieve very good performance, as the time of arrival of the next \gls{CDM} was within the predicted confidence range. On the other hand, Figure~\ref{fig:bad_prediction} serves as an illustrative instance of a less successful prediction, wherein the model falls in accurately determining the arrival time of the next \gls{CDM}. These contrasting scenarios highlight the model's ability and limitations, enriching our comprehensive evaluation of its predictive capacity.

\begin{figure*}[htb!]
    \centering
    \includegraphics[width=\textwidth]{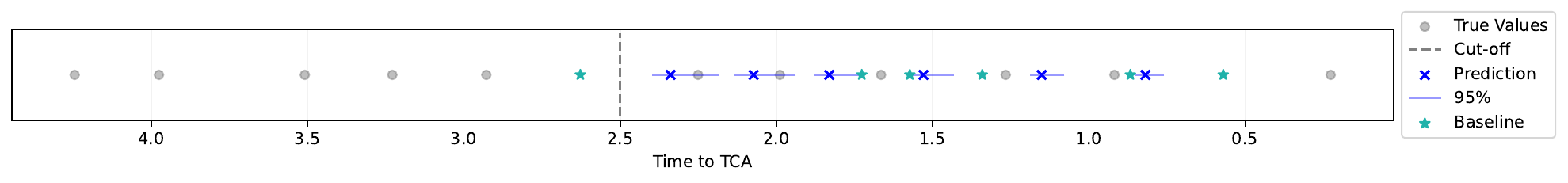}
    \caption{Example of full event prediction. The grey circles represent the real arrival times of CDMs, the stars indicate the prediction performed by the naïve Baseline model, and the blue crosses indicate the model predictions and the correspondent 95\% uncertainty bounds.}
    \label{fig:full_event_prediction}
\end{figure*}

In Figure \ref{fig:full_event_prediction}, we present the predictions of the subsequent \gls{CDM} arrival times for a randomly selected conjunction scenario. 

Upon analysis of the Figure~\ref{fig:full_event_prediction}, we observe that the \gls{NHPP} model demonstrates excellent performance in predicting the \gls{CDM} arrival times until the critical decision threshold of 2 days prior to the \gls{TCA}. Before this threshold, the model captures the temporal dynamics accurately, providing predictions that align closely with the actual \gls{CDM} arrival times. As expected, the uncertainty associated with each prediction decreases over time, reflecting the constant updating of the priors as new information becomes available.

However, beyond the critical decision threshold, we observe a decrease in the model's performance. Such a decrease in performance was, to some extent, anticipated, as the predictability of \gls{CDM} arrival times becomes more challenging as the \gls{TCA} is approached. Factors such as the inherent complexity of the conjunction scenario and the potential for last-minute adjustments in the screening processes can contribute to the dynamic nature of the \gls{CDM} issuance distribution near the \gls{TCA}. Nonetheless, the proposed \gls{NHPP} model still outperforms the baseline solutions in terms of accuracy and reliability near the critical decision threshold, providing valuable insights for collision avoidance decision-making.

To capture this changing distribution, i.e., multimodality of the data, and further improve predictions, future work could explore the application of a Poisson Mixture approach. By modelling the \gls{CDM} issuance as a mixture of Poisson distributions with varying parameters, this approach could effectively capture the evolving dynamics and uncertainty associated with the inter-\gls{CDM} times as the \gls{TCA} is approached.

\section{Conclusion}
We proposed a Bayesian non-homogeneous Poisson process model to predict the arrival time of \glspl{CDM}. The evaluation of the proposed solution revealed good performance, particularly evident up to the critical decision threshold, i.e., 2 days before the \gls{TCA}. However, we acknowledge that the model's predictive capacity decreases as the \gls{TCA} is approached. Such a phenomenon may arise due to the inherent complexity of the conjunction scenario and the potential for last-minute adjustments in the screening processes. Despite this limitation, our model outperformed both Baseline solutions presented, highlighting its potential to contribute to the development of automated collision avoidance systems and helping operators react timely but sparingly with satellite manoeuvres.

\section*{Acknowledgements}
This research was carried out under Project “Artificial Intelligence Fights Space Debris” Nº C626449889-0046305 co-funded by Recovery and Resilience Plan and NextGeneration EU Funds (www.recuperarportugal.gov.pt), and by NOVA LINCS (UIDB/04516/2020) with the financial support of FCT.IP.

%

\bibliographystyle{unsrtnat}

%



\end{document}